\pdfoutput=1

\documentclass[11pt]{article} %

\usepackage[final]{acl}

\usepackage{xcolor}

\definecolor{mine_red}{RGB}{239, 129, 131}
\definecolor{mine_blue}{RGB}{105, 158, 212}
\definecolor{mine_green}{RGB}{164, 217, 187}
\definecolor{mine_font}{RGB}{0, 128, 0}
\definecolor{light_mine_blue}{RGB}{145, 198, 252}
\definecolor{light_mine_red}{RGB}{255, 149, 151}

\definecolor{riptide}{RGB}{141,211,199}
\definecolor{pale_prim}{RGB}{255,255,179}
\definecolor{lavender_gray}{RGB}{190,186,218}
\definecolor{salmon}{RGB}{242,131,107}
\definecolor{seagull}{RGB}{128,177,211}
\definecolor{rajah}{RGB}{253,180,98}
\definecolor{yellow_green}{RGB}{198,222,119}
\definecolor{classic_rose}{RGB}{252,205,229}
\definecolor{feijoa}{RGB}{178,223,138}

\definecolor{cruise}{RGB}{179,226,205}
\definecolor{apricot}{RGB}{253,205,172}
\definecolor{periwinkle}{RGB}{203,213,232}
\definecolor{snow_flurry}{RGB}{230,245,201}
\definecolor{buttermilk}{RGB}{255,242,174}

\definecolor{sundown}{RGB}{249, 180, 181}
\definecolor{spindle}{RGB}{179,205,227}
\definecolor{tea_green}{RGB}{204,235,197}
\definecolor{languid_lavender}{RGB}{222,203,228}
\definecolor{champagne}{RGB}{254,217,166}
\definecolor{cream}{RGB}{255,255,204}

\definecolor{monte_carlo}{RGB}{135,204,194}
\definecolor{melon}{RGB}{254,191,181}
\definecolor{granny_smith_apple}{RGB}{150,214,150}
\definecolor{watusi}{RGB}{254,221,207}
\definecolor{see_green}{RGB}{161,228,195}

\definecolor{moss_green}{RGB}{170,216,176}
\definecolor{opal}{RGB}{164,207,190}

\definecolor{pale_turquoise}{RGB}{172,240,242}
\definecolor{Madang}{RGB}{190,235,159}
\definecolor{pixie_green}{RGB}{183,214,170}
\definecolor{coral_andy}{RGB}{243,204,205}
\definecolor{manhattan}{RGB}{226,180,125}
\definecolor{quartz}{RGB}{219,223,238}
\definecolor{spring_sun}{RGB}{242,243,195}
\definecolor{dairy_cream}{RGB}{254,226,189}
\definecolor{surf_crest}{RGB}{205,230,208}
\definecolor{french_pass}{RGB}{195,232,246}
\definecolor{cosmos}{RGB}{248,209,210}
\definecolor{portafino}{RGB}{245,237,160}
\definecolor{sail}{RGB}{163,205,235}
\definecolor{hint_green}{RGB}{226,246,209}

\definecolor{jet_stream}{RGB}{188, 214, 210}

\definecolor{azalea}{RGB}{251, 196, 196}
\definecolor{wewak}{RGB}{244, 143, 150}
\definecolor{bittersweet}{RGB}{255,111,105}
\definecolor{sunset_orange}{RGB}{242,89,75}
\definecolor{light_coral}{RGB}{244, 127, 123}
\definecolor{carnation}{RGB}{245, 80, 86}
\definecolor{flamingo}{RGB}{237, 88, 85}
\definecolor{carmine_pink}{RGB}{231, 76, 60}
\definecolor{deep_carmine_pink}{RGB}{236, 50, 67}
\definecolor{fire_engine_red}{RGB}{210,44,41}
\definecolor{amaranth}{RGB}{234,46,73}
\definecolor{ku_crimson}{RGB}{243, 0, 25}
\definecolor{fire_engine_red}{RGB}{206, 37, 51}
\definecolor{copper_rust}{RGB}{155, 64, 74}

\definecolor{chilean_fire}{RGB}{215, 87, 44}

\definecolor{japanese_laurel}{RGB}{53, 116, 40}

\definecolor{turmeric}{RGB}{211, 178, 76}
\definecolor{saffron}{RGB}{249,193,62}
\definecolor{my_sin}{RGB}{255, 176, 59}
\definecolor{tree_poppy}{RGB}{246, 154, 27}
\definecolor{jaffa}{RGB}{240, 131, 58}
\definecolor{crusta}{RGB}{254, 127, 44}
\definecolor{tahiti_gold}{RGB}{223, 102, 36}
\definecolor{outrageous_orange}{RGB}{255, 100, 45}
\definecolor{safety_orange}{RGB}{254, 106, 0}

\definecolor{turquoise}{RGB}{41,217,194}
\definecolor{puerto_rico}{RGB}{94, 194, 166}
\definecolor{mountain_meadow}{RGB}{0, 163, 136}
\definecolor{free_speech_aquamarine}{RGB}{0, 156, 114}
\definecolor{java}{RGB}{2,190,196}

\definecolor{matisse}{RGB}{25, 104, 167}
\definecolor{shakespeare}{RGB}{85, 154, 193}
\definecolor{mona_lisa}{RGB}{246,152,134}

\definecolor{bgc}{RGB}{245,245,245}
\definecolor{tuatara}{RGB}{67, 67, 67}
\definecolor{aluminum}{RGB}{153,153,153}
\definecolor{silver}{RGB}{191,191,191}
\definecolor{platinum}{RGB}{228,228,228}
\definecolor{mercury}{RGB}{230,230,230}
\definecolor{gallery}{RGB}{240,240,240}
\definecolor{athens_gray}{RGB}{236, 240, 241}
\definecolor{ship_gray}{RGB}{77,77,77}

\definecolor{early_dawn}{RGB}{252,243,218}
\definecolor{egg_shell}{RGB}{238, 234, 215}
\definecolor{midnight}{RGB}{0, 29, 50}
\definecolor{sundown}{RGB}{249, 180, 181}
\definecolor{sun_shade}{RGB}{255, 144, 68}
\definecolor{sushi}{RGB}{117, 168, 47}
\definecolor{tomato}{RGB}{255, 97, 56}
\definecolor{ice_cold}{RGB}{169,232,220}

\definecolor{jelly_bean}{RGB}{45, 126, 150}
\definecolor{celestial_blue}{RGB}{52, 152, 219}
\definecolor{curious_blue}{RGB}{41, 128, 185}
\definecolor{french_blue}{RGB}{0, 112, 182}
\definecolor{matisse}{RGB}{25, 104, 167}

\definecolor{biscay}{RGB}{44, 62, 80}

\definecolor{cosmic_latte}{RGB}{222, 247, 229}
\definecolor{chinook}{RGB}{163, 232, 178}
\definecolor{padua}{RGB}{121, 189, 143}
\definecolor{ocean_green}{RGB}{79, 176, 112}
\definecolor{pastel_green}{RGB}{107, 227, 135}
\definecolor{chateau_green}{RGB}{69, 191, 85}
\definecolor{RoyalBlue}{RGB}{69, 191, 85}
\definecolor{pigment_green}{RGB}{0, 175, 79}
\definecolor{fern}{RGB}{101,197,117}
\definecolor{killarney}{RGB}{56, 113, 66}
\definecolor{viridian}{RGB}{70, 137, 102}

\usepackage{booktabs}
\usepackage{colortbl}
\usepackage{adjustbox}
\usepackage{array}
\usepackage[group-separator={,}]{siunitx}
\usepackage{multirow}
\usepackage[inline]{enumitem}
\usepackage{bm}

\usepackage{times}
\usepackage{latexsym}

\usepackage[T1]{fontenc}

\usepackage[utf8]{inputenc}

\usepackage{microtype}

\usepackage{inconsolata}

\usepackage{graphicx}
\usepackage{subcaption}
\usepackage{amsmath}
\usepackage{makecell}   %

\renewcommand{\arraystretch}{1.12}

\title{Understanding the Collapse of LLMs in Model Editing}

\author{Wanli Yang$^{\spadesuit\heartsuit}$\hspace{0.7em} Fei Sun$^{\spadesuit}\footnotemark[2]$ \\ %
  \textbf{Jiajun Tan}$^{\spadesuit}$ \hspace{0.2em} \textbf{Xinyu Ma}$^{\clubsuit}$ \hspace{0.2em} \textbf{Du Su}$^{\spadesuit}$ \hspace{0.2em} \textbf{Dawei Yin}$^{\clubsuit}$ \hspace{0.2em} \textbf{Huawei Shen}$^{\spadesuit\heartsuit}$ \\
  $^{\spadesuit}$CAS Key Laboratory of AI Safety, Institute of Computing Technology, CAS\\
  $^{\heartsuit}$University of Chinese Academy of Sciences  \hspace{1.5em} $^{\clubsuit}$Baidu Inc. \\
 \tt{yangwanli24z@ict.ac.cn \,\,\, sunfei@ict.ac.cn}}

\begin{document}
\maketitle

\renewcommand*{\thefootnote}{\fnsymbol{footnote}}
\footnotetext[2]{Corresponding author.}
\renewcommand*{\thefootnote}{\arabic{footnote}}

\begin{abstract}

Despite significant progress in model editing methods, their application in real-world scenarios remains challenging as they often cause large language models (LLMs) to collapse.
Among them, ROME is particularly concerning, as it could disrupt LLMs with only a single edit.
In this paper, we study the root causes of such collapse.
Through extensive analysis, we identify two primary factors that contribute to the collapse:
\begin{enumerate*}[label=\roman*)]
    \item \textit{inconsistent handling of prefixed and unprefixed keys} in the parameter update equation may result in very small denominators, causing excessively large parameter updates;
    \item \textit{the subject of collapse cases is usually the first token}, whose unprefixed key distribution significantly differs from the prefixed key distribution in autoregressive transformers, causing the aforementioned issue to materialize.
\end{enumerate*}
To validate our findings, we propose a simple yet effective approach: uniformly using prefixed keys during editing phase and adding prefixes during testing phase to ensure the consistency between training and testing.
The experimental results show that the proposed solution can prevent model collapse while maintaining the effectiveness of the edits\footnote{Code and data are available at: \url{https://github.com/WanliYoung/Collapse-in-Model-Editing}.}.

\end{abstract}

\section{Introduction}

Recent works \cite{yang2024butterfly, gupta2024model, gu2024model} have revealed that model editing \cite{zhang2024comprehensive} poses significant risks of compromising the capabilities of large language models (LLMs).
Among them, Rank-One Model Editing (ROME) \cite{meng2022locating}, a cutting-edge method, has been found to cause model collapse with just a single edit \cite{yang2024butterfly}.
In this paper, we aim to study the underlying causes behind this phenomenon.

Intuitively, for a knowledge tuple (subject, relation, object), ROME takes a prompt constructed from the subject and relation as input and models the knowlege in a key-value format.
Here, the key is a vector representation of the subject within the prompt, and the value is a vector representation capable of yielding the target object, obtained by transforming the key through a transformation matrix.
To insert a new fact about a subject, ROME adjusts the transformation matrix to match the key of the subject with the value of the new fact, as described in Eq.~\ref{Eqn_update}.

To uncover the underlying causes of ROME's collapse, we investigate the differences in parameter update process of ROME between \textit{collapse cases} (i.e., samples that induce collapse) and \textit{normal cases} (i.e., samples that do not).
The results reveal that the collapse directly stems from the anomalously small denominator within the parameter update equation (Eq.~\ref{Eqn_update}). 
This anomaly originates from the irregular implementation of the keys in the denominator, where one is derived by prepending varying prefixes to the subject to simulate diverse contexts (termed \textit{prefixed key}), while the other is obtained directly from the original subject without any prefix (termed \textit{unprefixed key}).
This issue has also been independently identified by \citet{gupta2024rebuilding} concurrently.
However, it is still unclear why the irregular implementation only fails in collapse cases.

To answer this question, we examine the distribution of elements in the denominator. %
It reveals that, in collapse cases, the distribution of the unprefixed keys exhibits significant difference from the prefixed keys.
This leads to an exceptionally small denominator in the update equation, which in turn causes the model to collapse.

\begin{figure*}[!ht]
    \centering
    \includegraphics[width=\linewidth]{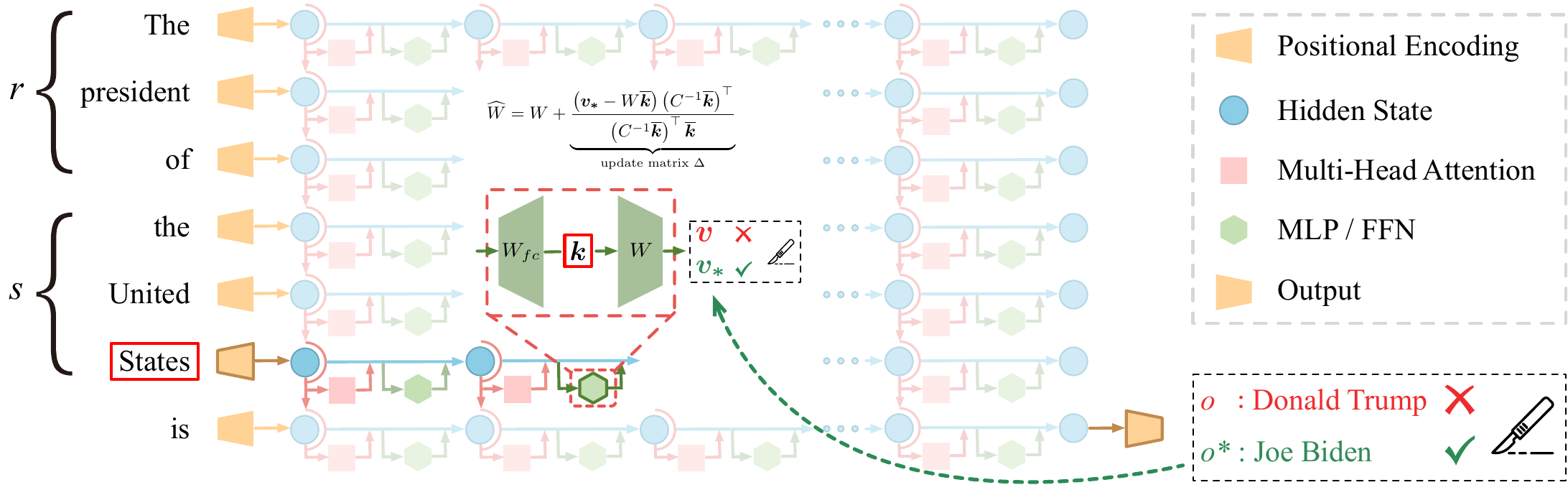}
    \caption{To update ``the president of the United States'' from ``Donald Trump'' to ``Joe Biden'', ROME locates the knowledge into the MLP module within a specific transformer block using the Causal Tracing mechanism. 
    It then adjusts the second layer of MLP (i.e., weight matrix $W$) to change the value $\bm{v}$ for the key $\bm{k}$ that represents the subject ``the United States'' to a new value $\bm{v}_*$, thereby inducing the LLMs to predict the target object ``Joe Biden''.}
    \label{fig_rome}
\end{figure*}

To elucidate the anomalous behavior observed in the collapse cases, we conduct an analysis starting from their characteristics.
The collapse cases of both GPT-2-XL \cite{radford2019language} and GPT-J \cite{gpt-j} exhibit a consistent pattern: \textit{the subjects in nearly all of these instances correspond to the first tokens within their respective prompts}. %
Furthermore, we discover that \textit{the representation distribution of the first tokens markedly diverges from that of the subsequent tokens in these autoregressive models}.
These two factors, working in concert, lead to the anomalous distribution of unprefixed keys in collapse cases.

To validate our findings, we propose unifying all keys as prefixed during editing to prevent model collapse. 
To ensure consistency with the editing process, when using the edited model, we prefix a random text for instances where subjects are in the first token.
Experiments validate that our proposed method effectively prevents model collapse while ensuring the success of edits.

Our main contributions are as follows:
\begin{itemize} [itemsep=3pt, topsep=3pt, parsep=3pt]
    \item Comprehensive analysis that identifies two factors behind ROME's collapse: i) inconsistent implementation of key vectors; ii) anomalous distribution of first token representations.
    \item A straightforward solution to prevent collapse while maintaining editing efficacy. %
\end{itemize}

\section{Background}
\label{Sec_Back}

ROME \cite{meng2022locating} hypothesizes that the MLP modules in the Transformer architecture \cite{vaswani2017attention} can be modeled as a linear key-value associative memory.
Under the hypothesis in ROME, a knowledge triplet $(s, r, o)$ corresponds to a key-value pair $(\bm{k}, \bm{v})$, where $\bm{k}$ represents the subject $s$, and $\bm{v}$ encodes the property $(r, o)$ for $s$. 
The entire knowledge within a model can thus be represented as a set of key vectors $K = [\bm{k}_1, \dots, \bm{k}_n]$ and value vectors $V = [\bm{v}_1, \dots, \bm{v}_n]$.
A linear operation $W$ matches keys to values by solving $W K \approx V$.

In practice, for an input prompt $\mathtt{p}(s, r)$, the recall of the target object $o$ mainly occurs within a two-layer MLP in a specific transformer block identified by the Causal Tracing mechanism \cite{meng2022locating}.
Specifically, output of the first layer for the subject $s$ forms a key $\bm{k}$, and the second layer (parameterized with $W$) retrieves an associated value $\bm{v}$ based on this key $\bm{k}$, ultimately inducing the LLMs to predict the target object $o$.

In this context, to replace the current knowledge  $(s, r, o)$ with a new knowledge tuple $t^* =(s, r, o^*)$, we need to find the corresponding key $\bm{k}$ and the new value $\bm{v}_*$. 
To simulate various contexts for generalization, ROME assigns $\bm{k}$ as an average vector $\overline{\bm{k}}$ derived from subject $s$ with a small set of $N$ randomly sampled prefixes:
\begin{equation}
    \overline{\bm{k}} = \frac{1}{N} \sum_{i=1}^N \mathcal{K}\left(x_i \oplus s\right) \label{Eqn_prefix_key} %
\end{equation}
where $\mathcal{K}$ is the output of the first MLP layer in transformer block, $x_i$ is the prefixes, and $\oplus$ is string concatenation operator.

\label{edited_prompt}
To illustrate the selection of $\bm{v}_*$, we take the subject $s{=}$ ``\textit{United States}'' and relation $r{=}$ ``\textit{president of}'' as an example.
A specifically designed loss function is utilized to optimize $\bm{v}_*$ so that it can produce $o^*=$ ``\textit{Joe Biden}'' when given the prompt $\mathtt{p}(s, r)=$ ``\textit{The president of the United States is}''.

With the computed $(\overline{\bm{k}}, \bm{v}_*)$, ROME finds optimal $\widehat{W}$ by solving the following problem: %
\begin{equation}
   \underset{\widehat{W}}{\arg\min} \|\widehat{W} K-V\| \ \text{ subject to } \widehat{W} \overline{\bm{k}}=\bm{v}_*
\end{equation}
It has the following closed-form solution:
\begin{equation}
\label{Eqn_update}
    \widehat{W} = W + \underbrace{\frac{\left(\bm{v}_* - W \overline{\bm{k}}\right) \left(C^{-1} \overline{\bm{k}}\right)^\top}{\left(C^{-1} \overline{\bm{k}}\right)^\top \overline{\bm{k}}}}_{\text{update matrix } \Delta}
\end{equation}
where $W$ denotes the weight matrix of the second layer in the MLP before editing, $\widehat{W}$ denotes the weight matrix after editing, and $C{=}KK^\top$ is a pre-cached constant. 

The complete editing process of ROME is illustrated in Figure~\ref{fig_rome}.
Interested readers are directed to \citet{meng2022locating} for a detailed introduction.

\section{Why Does ROME Cause Collapse?}

Previous studies \cite{yang2024butterfly, gupta2024model} have revealed that a single edit of ROME can induce LLMs to collapse.
To further analyze the cause, we investigate the differences in parameter updates between samples that induce collapse and those do not.
For this purpose, we introduce two distinct subsets: i) \textit{collapse cases}, using the \textit{HardCF} set built by \citet{yang2024butterfly}, which includes collapse cases on GPT-2-XL, GPT-J, and Llama2-7b from the COUNTERFACT dataset \cite{meng2022locating}; and ii) \textit{normal cases}, comprising 1000 random samples from the remaining part of COUNTERFACT.

\subsection{Inconsistent Keys in Editing}

\begin{table}[t]
    \centering
    \renewcommand{\arraystretch}{1.15}
    \begin{adjustbox}{max width=\linewidth} 
    \begin{tabular}{llrrr}
        \toprule
        Component & Cases & GPT-2-XL & GPT-J & Llama2-7b \\
        \midrule
        numerator: & collapse & \num{168.55} & \num{140.27} & \num{4.57} \\
        $\left(\bm{v}_*-W\overline{\bm{k}}\right) \left(C^{-1} \overline{\bm{k}}\right)^\top$& normal & \num{79.91} & \num{88.69} & \num{16.52} \\

        \midrule
        denominator: & \cellcolor{mercury}collapse & \cellcolor{mercury}
\num{0.04} & \cellcolor{mercury}\num{0.04} & \cellcolor{mercury}\num{0.01} \\
        $\left(C^{-1} \overline{\bm{k}}\right)^\top \overline{\bm{k}}$ & normal & \num{9.60} & \num{12.78} & \num{2.63} \\

        \bottomrule
    \end{tabular}
    \end{adjustbox}
    \caption{Average norm of the numerator and average absolute value of the denominator in ROME's update matrix $\Delta$ across various LLMs for different sets of cases.}
    \label{tab_DivF}
\end{table}

Existing work \cite{yang2024butterfly} has found that collapse is caused by the values of update matrix $\Delta$ in Eq.~\ref{Eqn_update} being excessively large.
For fine-grained analysis, we split $\Delta$ into \textit{numerator} (a matrix) and \textit{denominator} (a scalar), and then apply single edits to analyze the intermediate values for parameter updating in different cases.
As illustrated in Table~\ref{tab_DivF}, the denominators of collapse cases are two orders of magnitude smaller than those of normal cases, while the numerators do not show significant differences. 
This disparity directly results in the exceptionally large $\Delta$ of collapse cases. %

These results guide our focus to the key $\overline{\bm{k}}$ in the denominator $(C^{-1} \overline{\bm{k}})^\top \overline{\bm{k}}$, given that the matrix $C$ is a constant for both collapse cases and normal cases.
We revisited the official implementation of ROME and identified that \textbf{different variants of $\overline{\bm{k}}$ are used}.
Specifically, only $\overline{\bm{k}}$ within $(C^{-1} \overline{\bm{k}})^\top$ is the prefixed key as in Eq.~\ref{Eqn_prefix_key}. %
In contrast, \textbf{$\overline{\bm{k}}$ in other positions is unprefixed}, utilizing a representation over the subject $s$ without any prefix, denoted as $\bm{k}^{u} = \mathcal{K}\left(s\right)$.
However, ideally, all $\overline{\bm{k}}$ in Eq.~\ref{Eqn_update} should be the same, i.e., the average representation derived from a set of prefixed subjects as in Eq.~\ref{Eqn_prefix_key}.

\label{C-ROME}
To verify if this inconsistency of keys is responsible for the collapse, we substitute all $\bm{k}^{u}$ with $\overline{\bm{k}}$ in the implementation. 
The aligned implementation is referred to as \textit{Consistent-ROME}, \textit{C-ROME} for short.
We evaluate the different implementations on collapse and normal cases using perplexity on the ME-PPL$_{50}$ dataset, whose effectiveness has been validated by \citet{yang2024butterfly}. 
According to Table~\ref{tab_fix}, C-ROME with aligned implementation of $\overline{\bm{k}}$ does not significantly alter the edited models, avoiding the sharp increase in perplexity seen with ROME. 
This demonstrates that such inconsistency of $\overline{\bm{k}}$ in the update matrix $\Delta$ is a primary factor behind ROME-induced model collapse.

\begin{table}[t]
    \centering
    \renewcommand{\arraystretch}{1.15}
    \begin{adjustbox}{max width=\linewidth} 
    \begin{tabular}{llrrr}
        \toprule
        Method & Cases & GPT-2-XL & GPT-J & Llama2-7b \\
        \midrule
        Original &  & \num{68.77} & \num{49.04} & \num{33.18} \\
        \midrule
        \multirow{2}{*}{ROME} & collapse & \num{26084.66} & \num{25909.24} & \num{10574.76} \\
         & normal & \num{74.32} & \num{50.77} & \num{36.68} \\
        \midrule
        \multirow{2}{*}{C-ROME} & collapse & \num{70.71} & \num{51.77} & \num{33.20} \\
         & normal & \num{70.28} & \num{50.57} & \num{33.55} \\
        \bottomrule
    \end{tabular}
    \end{adjustbox}
    \caption{The maximum ME-PPL$_{50}$ perplexity of models edited by different implementations of ROME for their collapse cases and normal cases, with their original models' perplexity for comparison.}
    \label{tab_fix}
\end{table}

\subsection{Anomalous Key Distribution for Collapse}

While unifying the keys as $\overline{\bm{k}}$ can prevent model collapse, it remains unclear \textit{why inconsistent keys only encounter issues in collapse cases}.

\begin{figure}[t]
  \centering

  \begin{subfigure}[t]{0.48\columnwidth}
    \includegraphics[width=\linewidth]{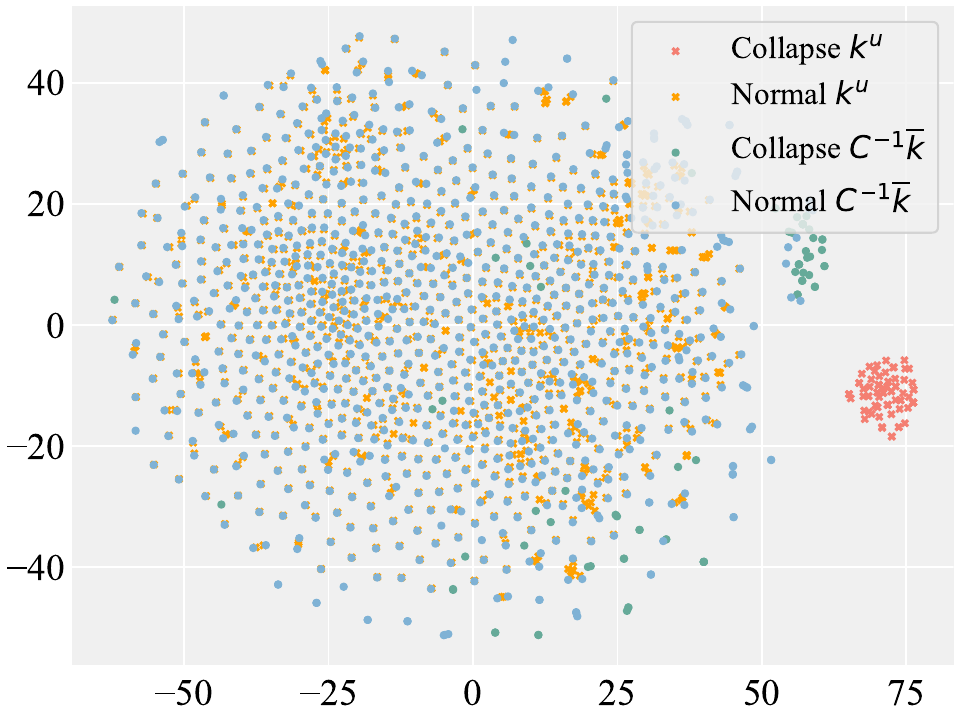}
    \captionsetup{skip=3pt}
    \caption{}
    \label{subfig_keys-keyc}
  \end{subfigure}
  \hfill %
  \begin{subfigure}[t]{0.48\columnwidth}
    \includegraphics[width=\linewidth]{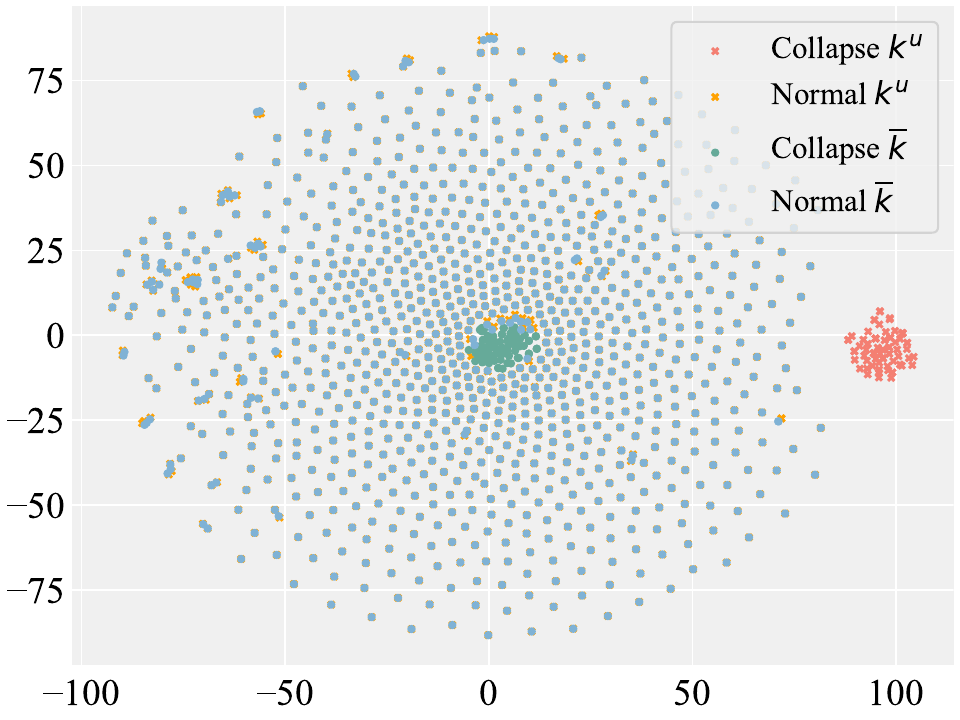}
    \captionsetup{skip=3pt}
    \caption{}
    \label{subfig_key-vectors}
  \end{subfigure}
  \caption{t-SNE visualization of (a) elements in the denominator; (b) different implementation of key vectors.}
  \label{fig_dist}
\end{figure}

To enhance intuitive understanding, we analyze the spatial distribution of $C^{-1} \overline{\bm{k}}$ and $\bm{k}^{u}$ in the denominator for different cases by projecting them into a two-dimensional space using t-SNE~\cite{van2008visualizing}. %
Taking the results of GPT-2-XL in Figure~\ref{subfig_keys-keyc} as an example, in normal cases, the distribution of $C^{-1} \overline{\bm{k}}$ and $\bm{k}^{u}$ show no significant differences. 
However, a noticeable divergence in the distribution occurs in collapse cases, explaining the exceptionally small denominators.

Considering that $C$ is a constant, the differences between normal and collapse cases should arise from the variations in the prefixed key $\overline{\bm{k}}$ and the unprefixed key $\bm{k}^{u}$.
Figure~\ref{subfig_key-vectors} clearly illustrates that the distribution of $\bm{k}^{u}$ in collapse cases significantly diverge from those of $\overline{\bm{k}}$.
This confirms that in collapse cases, the significant differences between $\overline{\bm{k}}$ and $\bm{k}^{u}$ result in a particularly small denominator in the update matrix, which in turn leads to the collapse of the edited model.
Similar phenomena are also observed in other LLMs, detailed in \S~\ref{apd_dist_other}.

\subsection{Special Role of the First Token}

To elucidate the anomalous distribution of $\bm{k}^{u}$ in collapse cases, we focus our analysis on their characteristics. 
A common pattern is observed in the collapse cases for both GPT-2-XL and GPT-J: \textit{in almost all instances, the subjects consist of a single word, which is encoded as a single token and positioned at the beginning of the input prompt $\mathtt{p}(s, r)$}\footnote{The only exception involves few instances with subjects like ``Jackson Jackson'' in the collapse cases of GPT-J.}. 
Therefore, the unprefixed key $\bm{k}^{u}$ for a collapse case is the intermediate representation within the MLP layer of the first token in the input.
This inspires us to investigate whether the anomalous distribution of $\bm{k}^{u}$ in collapse cases can be attributed to their position as the first tokens in the prompts.

\begin{figure}[t]
  \centering

  \begin{subfigure}[t]{0.48\columnwidth}
    \includegraphics[width=\linewidth]{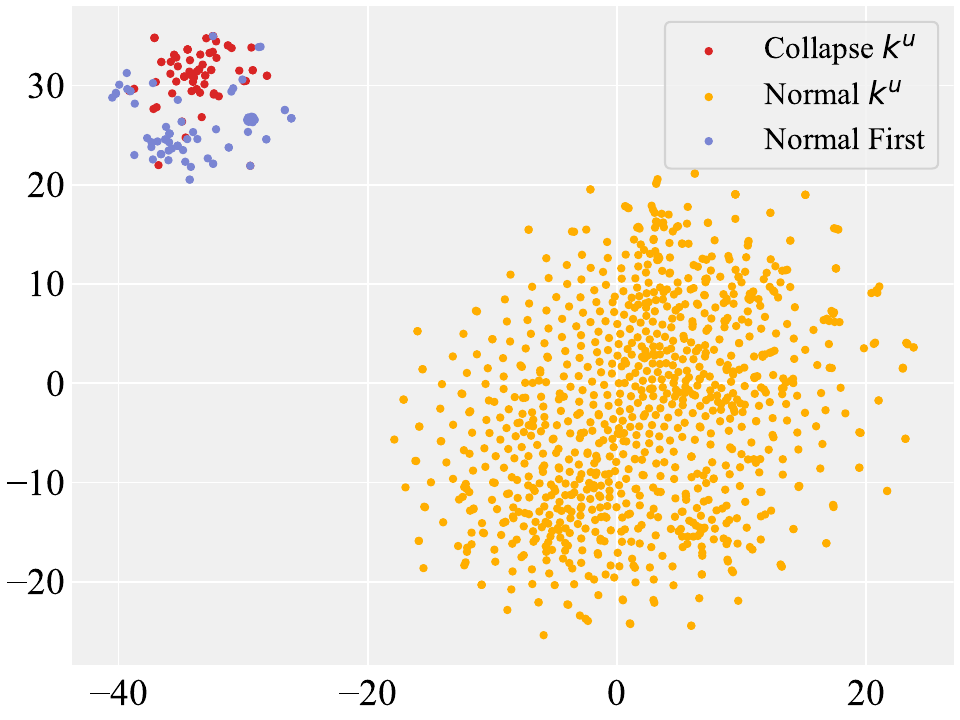}
    \captionsetup{skip=3pt}
    \caption{}
    \label{subfig_norm_sentence}
  \end{subfigure}
  \hfill %
  \begin{subfigure}[t]{0.48\columnwidth}
    \includegraphics[width=\linewidth]{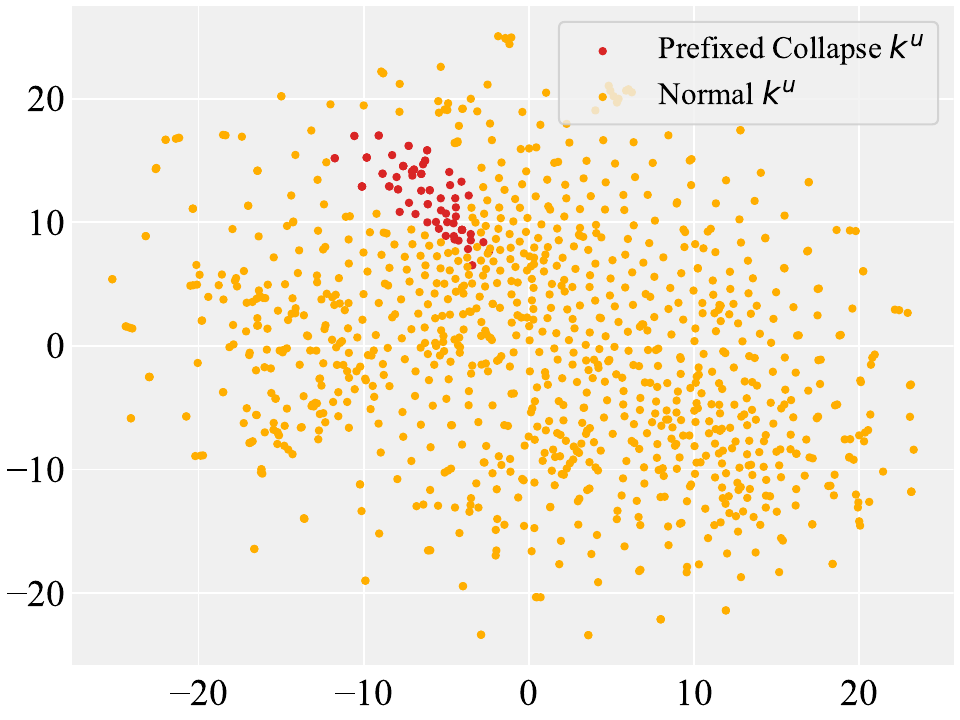}
    \captionsetup{skip=3pt}
    \caption{}
    \label{subfig_prefixed_collapse}
  \end{subfigure}
  \caption{t-SNE visualization of representation distributions of (a) the first token in randomly sampled normal prompts; (b) $\bm{k}^{u}$ in prefixed collapse prompts.}
  \label{fig_sentence}
\end{figure}

To explore this, we first examined the representation distribution of the first tokens in the prompts for normal cases.
The results presented in Figure~\ref{subfig_norm_sentence} indicate that, within GPT-2-XL, the first tokens of normal cases consistently exhibit an abnormal distribution similar to that of $\bm{k}^{u}$ in collapse cases. 
From an opposing perspective, to verify whether artificially shifting the $\bm{k}^{u}$ in collapse cases away from the first position would eliminate the anomaly in distribution, we prefixed the prompts of collapse cases with randomly sampled texts.
This adjustment results in their distribution aligning with that of normal cases, as illustrated in Figure~\ref{subfig_prefixed_collapse}.
These findings suggest that the anomalous distribution of $\bm{k}^{u}$ for collapse cases in ROME is not related to the editing process.
Instead, it is due to the unique pattern of their subjects encountering the special distribution of the first token in GPT-2-XL and GPT-J models.

It is important to note that Llama2-7b \cite{touvron2023llama}, Mistral-7b \cite{jiang2023mistral7b}, and Llama3-8b \cite{llama3} avoid collapse in such cases due to their tokenizers additionally incorporating a special token, e.g., \texttt{<s>}, at the beginning of the input, which shifts the subject from being the first token.
In fact, we found they also succumb to collapse when the special token is removed, with results detailed in Appendix~\ref{apd_del_llama}.

\textbf{\textit{Analysis.}} 
To elucidate the underlying reasons for the anomalous distribution of the first token in autoregressive language models, we explored two potential factors as follows.

Firstly, we speculate that this phenomenon may arise from the inherent nature of autoregressive models, where the first token cannot interact with any other token except itself.
As a counterexample with non-autoregressive architecture, the representation distribution of the first token in T5-3B encoder \cite{2020t5} does not differ from that of subsequent tokens.
This may be attributed to the bidirectional attention in the encoder, which enables interactions between the first token and subsequent tokens.
A detailed analysis is presented in Appendix~\ref{apd_t5}.

Secondly, considering the specificity of the first token may originate from its position embedding, we verify it from two aspects.
For \textit{collapse cases} where the subjects are the first tokens, setting the position embedding of the first token as that of the second token can not completely eliminate collapse. 
While for \textit{normal cases} where the subjects are the second tokens, replicating the position embedding of the first token onto the second token does not consistently lead to collapse.
These findings suggest that while position embedding plays a role, it is not the only determining factor.
The detailed investigation is provided in Appendix~\ref{apd_position_embd}.

Additionally, we observed that in GPT-2-XL and GPT-J, the representations of the first tokens rapidly become significantly more concentrated than those of subsequent tokens as the layers progress.
However, this phenomenon does not appear in Llama2-7b, Mistral-7b, and Llama3-8b. 
A detailed investigation is presented in Appendix~\ref{apd_collapse_1token}.

\begin{table}[t]
    \centering
    \begin{adjustbox}{max width=\linewidth} 
    \begin{tabular}{lrrrr}
        \toprule
        Model & GPT-2-XL & GPT-J & Mistral-7b & Llama3-8b \\
        \midrule
        Ori PPL & \num{68.39} & \num{50.34} & \num{51.75} & \num{41.67} \\
        \midrule
        Max PPL & \num{68.91} & \num{50.59} & \num{52.19} & \num{43.98} \\
        \bottomrule
    \end{tabular}
    \end{adjustbox}
    \caption{The maximum perplexity for various LLMs edited by ROME on the collapse cases of Llama2-7b, with their original perplexity for comparison.}
    \label{tab_no_collapse}
\end{table}

Regarding the collapse cases of Llama2-7b, we found that the subjects of them terminate with a period ``.''.
It is worth noting that, such cases are extremely rare, amounting to just \num{21} out of \num{21919} samples in the COUNTERFACT dataset. 
Furthermore, they do not induce model collapse in various other models, including GPT-2-XL, GPT-J, Mistral-7b and Llama3-8b (the successor of Llama2-7b), as shown in Table~\ref{tab_no_collapse}.
Consequently, we have decided not to pursue an exhaustive investigation of this isolated phenomenon.

\begin{table}[t]
    \centering
    \begin{adjustbox}{max width=0.96\linewidth} 
    \begin{tabular}{lrrr}
        \toprule
        Model & efficacy & generalization & locality \\
        \midrule
        GPT-2-XL & \num{5.19}\% & \num{14.29}\% & \num{97.40}\% \\
        \midrule
        GPT-J & \num{30.59}\% & \num{30.77}\% & \num{82.35}\% \\
        \midrule
        Llama2-7b & \num{18.65}\% & \num{12.70}\% & \num{100}\% \\
        \bottomrule
    \end{tabular}
    \end{adjustbox}
    \caption{Performance of C-ROME on various LLMs for corresponding collapse cases.
    Notably, the efficacy in normal cases typically exceeds 90\%.}
    \label{tab_metric}
\end{table}

\section{A Simple Solution to Avoid Collapse}
Having identified the reasons for ROME's collapse, it is crucial to provide a solution to prevent these problems.
C-ROME introduced in \S~\ref{C-ROME} can effectively keep the stability of edited models, but Table~\ref{tab_metric} reveals that it fails to successfully integrate target knowledge into the model, as evidenced by its low \textit{efficacy} and \textit{generalization} \cite{yao-etal-2023-editing} metrics on collapse cases.
This failure arises from the inconsistency of C-ROME between editing and testing. 
Specifically, C-ROME employs prefixed keys $\overline{\bm{k}}$ only when editing, while during testing, the prompts used to evaluate efficacy adopt unprefixed keys $\bm{k}^{u}$, which significantly differ from $\overline{\bm{k}}$.
This inconsistency results in an inability to obtain the appropriate target value vector corresponding to the key of collapse cases, finally leading to low efficacy of editing.

To address this issue, we propose a straightforward solution, which appends a random prefix, drawn from those utilized in the editing process, to the prompt of collapse cases during the testing phase.
The results in Table~\ref{tab_prefix_rome} demonstrate that this method significantly improves the efficacy for GPT-2-XL, GPT-J, and Llama2-7b, albeit with a relatively limited improvement of generalization.

\begin{table}[t]
    \centering
    \begin{adjustbox}{max width=\linewidth} 
    \begin{tabular}{llrrr}
        \toprule
        Model & Cases & efficacy & generalization & locality \\
        \midrule
        \multirow{2}{*}{GPT-2-XL} & collapse & \num{100}\% & \num{16.88}\% & \num{100}\% \\
         & normal & \num{96.16}\% & \num{41.88}\% & \num{97.34}\% \\
        \midrule
        \multirow{2}{*}{GPT-J} & collapse & \num{100}\% & \num{32.94}\% & \num{89.41}\% \\
         & normal & \num{99.77}\% & \num{50.00}\% & \num{95.61}\% \\
        \midrule
        \multirow{2}{*}{Llama2-7b} & collapse & \num{91.27}\% & \num{29.37}\% & \num{100}\% \\
         & normal & \num{91.95}\% & \num{46.73}\% & \num{97.56}\% \\
        \bottomrule
    \end{tabular}
    \end{adjustbox}
    \caption{Performance of C-ROME, enhanced by prefixing random texts to the prompts of collapse cases during testing, across various LLMs on both collapse cases and the remaining data within COUNTERFACT.}
    \label{tab_prefix_rome}
\end{table}

\section{Conclusion and Future Work}

In this paper, we thoroughly investigate the underlying causes of LLMs collapse triggered by a single edit of ROME.
Our extensive experiments demonstrate that such collapse arises from two aspects:
\begin{enumerate*}[label=\roman*)]
\item irregularities in the official implementation of ROME, which employs two types of keys in parameter updating;
\item anomalous representation distribution of the first token in autoregressive models.
\end{enumerate*}
Consequently, we propose a straightforward and simple method to address the model collapse issue of ROME, and validate its effectiveness with extensive experiments 

For future research, we intend to investigate the root causes of model collapse in sequential editing and to devise more robust editing methods that ensure the stability of the edited model and superior editing performance across various scenarios.

\clearpage
\section*{Limitations}

We acknowledge following limitations of our work:

\begin{itemize}[itemsep=0pt, topsep=0pt, parsep=0pt]
\item The analysis in this paper primarily focuses on GPT-2-XL and GPT-J. 
Regarding Llama2-7b, which exhibits a unique pattern of collapse cases, our solution successfully prevents its collapse. 
However, the specific characteristics of its collapse cases remain unknown.
\item Due to space limitations, we have left an in-depth investigation into the anomalous representation distribution of the first token in autoregressive models for future research. This anomaly represents a broader issue that requires further exploration.
\item This paper focuses on the root causes of model collapse triggered by a single edit of ROME.
The collapse resulting from the cumulative effects of sequential editing, a phenomenon observed in existing works, is beyond the scope of this paper and is reserved for future work.
\end{itemize}

\section*{Acknowledgements}
This work was supported by the National Key R\&D Program of China (2022YFB3103700, 2022YFB3103704), the Strategic Priority Research Program of the Chinese Academy of Sciences (No. XDB0680101), and the Innovation Funding of ICT, CAS (E361120).

\bibliography{custom}

\clearpage
\appendix
\section{Appendix}
\label{sec:appendix}

\subsection{Distribution of Keys in Other LLMs}
\label{apd_dist_other}

\begin{figure}[t]
  \centering

  \begin{subfigure}[t]{0.49\columnwidth}
    \includegraphics[width=\linewidth]{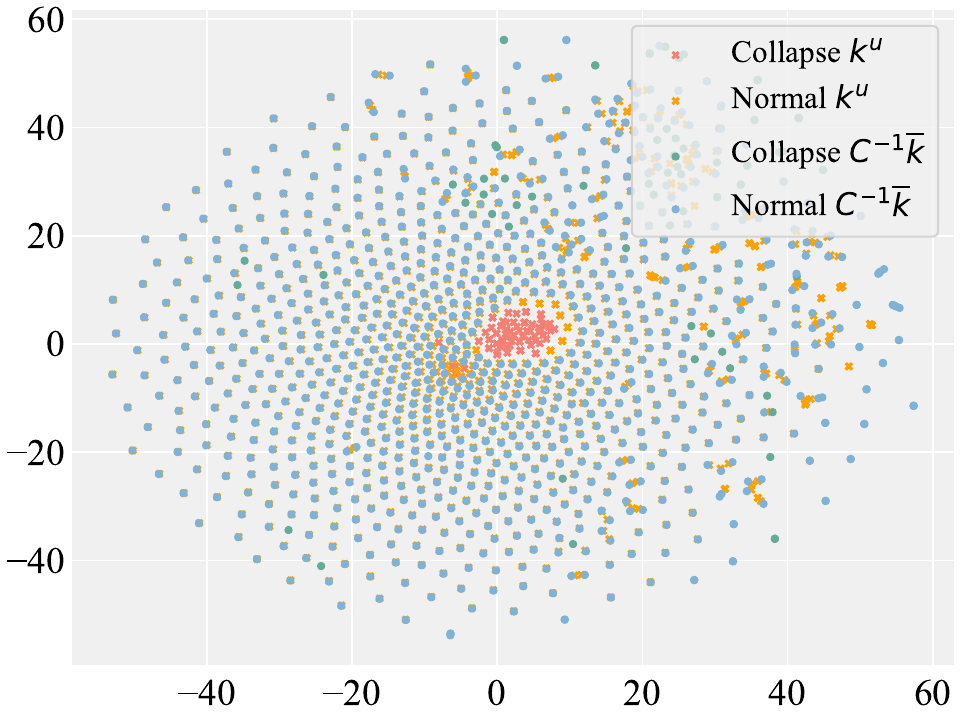}
    \caption{}
    \label{subfig_keys-keyc_gptj}
  \end{subfigure}
  \begin{subfigure}[t]{0.49\columnwidth}
    \includegraphics[width=\linewidth]{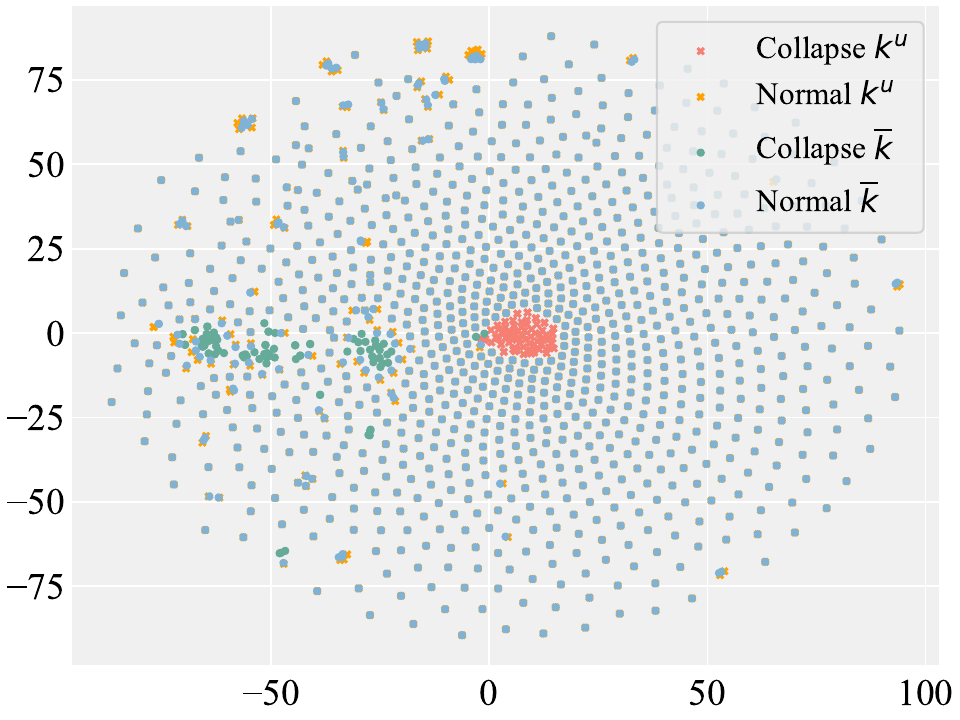}
    \caption{}
    \label{subfig_key-vectors_gptj}
  \end{subfigure}

  \caption{t-SNE visualization of (a) elements in the denominator; (b) different implementation of key vectors for GPT-J.}
  \label{fig_dist_gptj}
\end{figure}

\begin{figure}[t]
  \centering

  \begin{subfigure}[t]{0.49\columnwidth}
    \includegraphics[width=\linewidth]{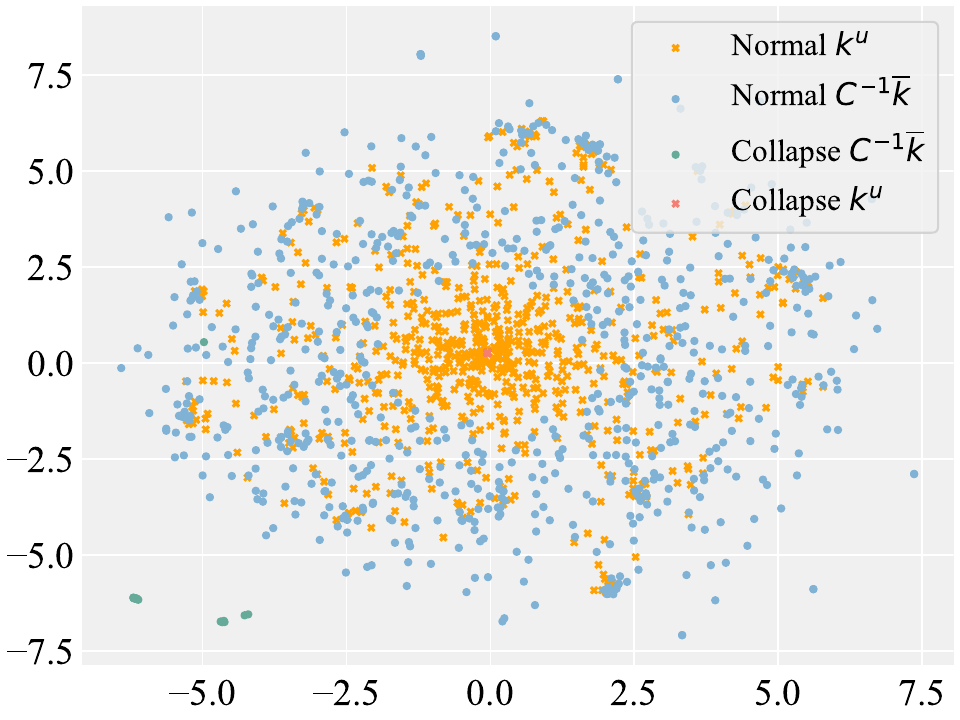}
    \caption{}
    \label{subfig_keys-keyc_llama}
  \end{subfigure}
  \begin{subfigure}[t]{0.49\columnwidth}
    \includegraphics[width=\linewidth]{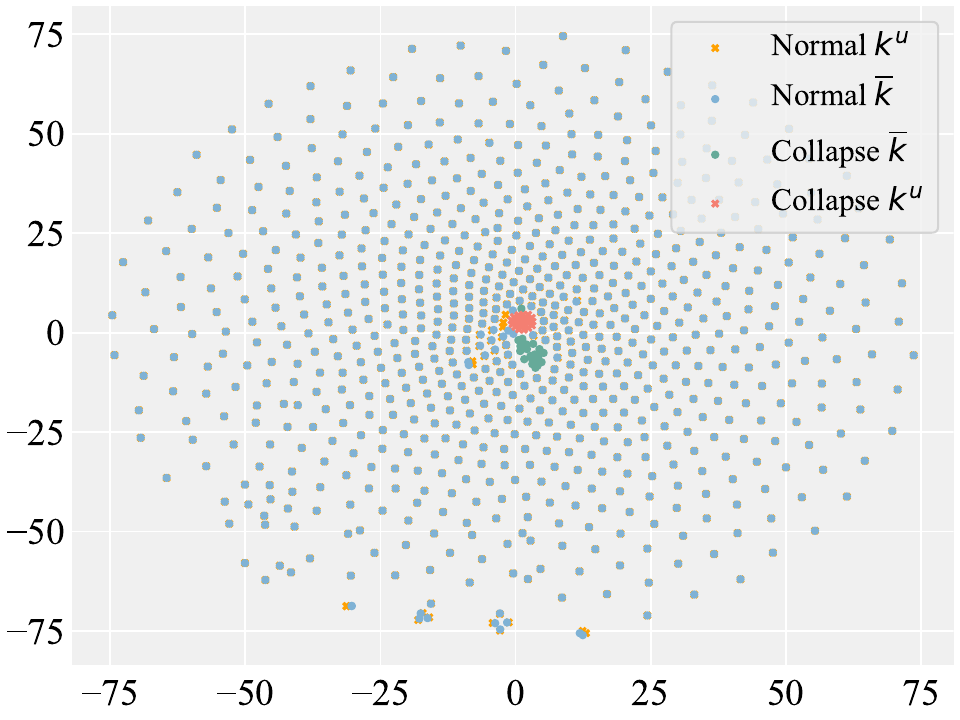}
    \caption{}
    \label{subfig_key-vectors_llama}
  \end{subfigure}

  \caption{t-SNE visualization of (a) elements in the denominator; (b) different implementation of key vectors for Llama2-7b.}
  \label{fig_dist_llama}
\end{figure}

The distribution of $C^{-1} \overline{\bm{k}}$ and $\bm{k}^{u}$ for collapse and normal cases of GPT-J in two-dimensional space is shown in Figure~\ref{subfig_keys-keyc_gptj}, demonstrating a significant difference between the distributions of these two elements in collapse cases.
The results for $\overline{\bm{k}}$ and $\bm{k}^{u}$ is depicted in Figure~\ref{subfig_key-vectors_gptj}, revealing similar disparities.
The corresponding results for Llama2-7b are provided in Figure~\ref{subfig_keys-keyc_llama} and Figure~\ref{subfig_key-vectors_llama}, showing consistent phenomena.

\subsection{Results without Prepended Token}
\label{apd_del_llama}

\begin{figure}[t]
  \centering

  \begin{subfigure}[t]{0.49\columnwidth}
    \includegraphics[width=\linewidth]{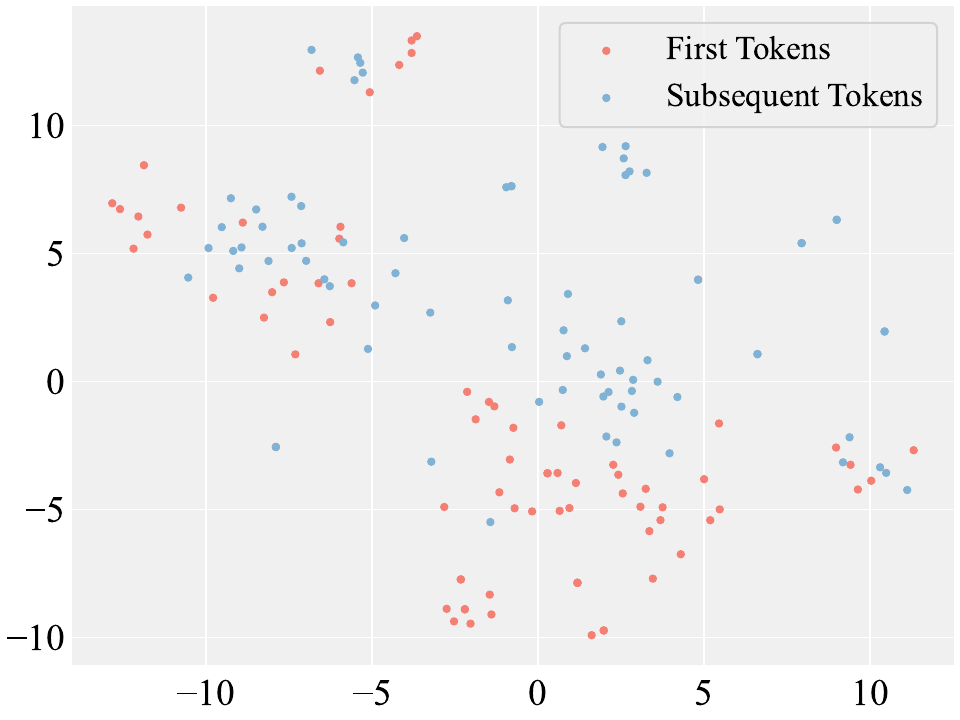}
    \captionsetup{skip=2pt}
    \caption{layer 1}
  \end{subfigure}
  \begin{subfigure}[t]{0.49\columnwidth}
    \includegraphics[width=\linewidth]{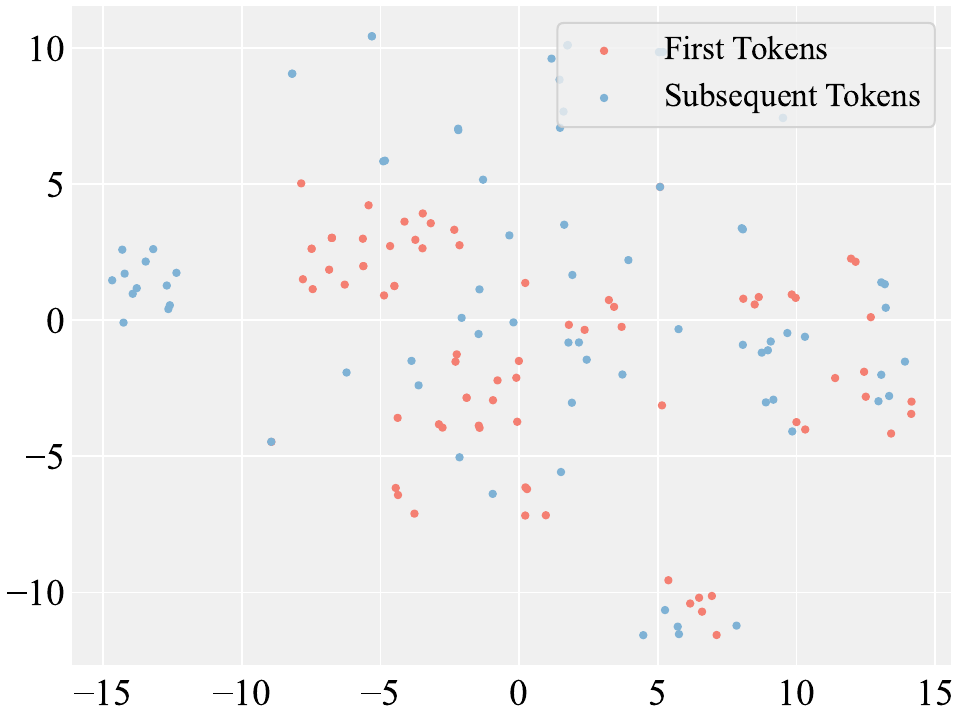}
    \captionsetup{skip=2pt}
    \caption{layer 5}
  \end{subfigure}

  \vspace{5pt}

  \begin{subfigure}[t]{0.49\columnwidth}
    \includegraphics[width=\linewidth]{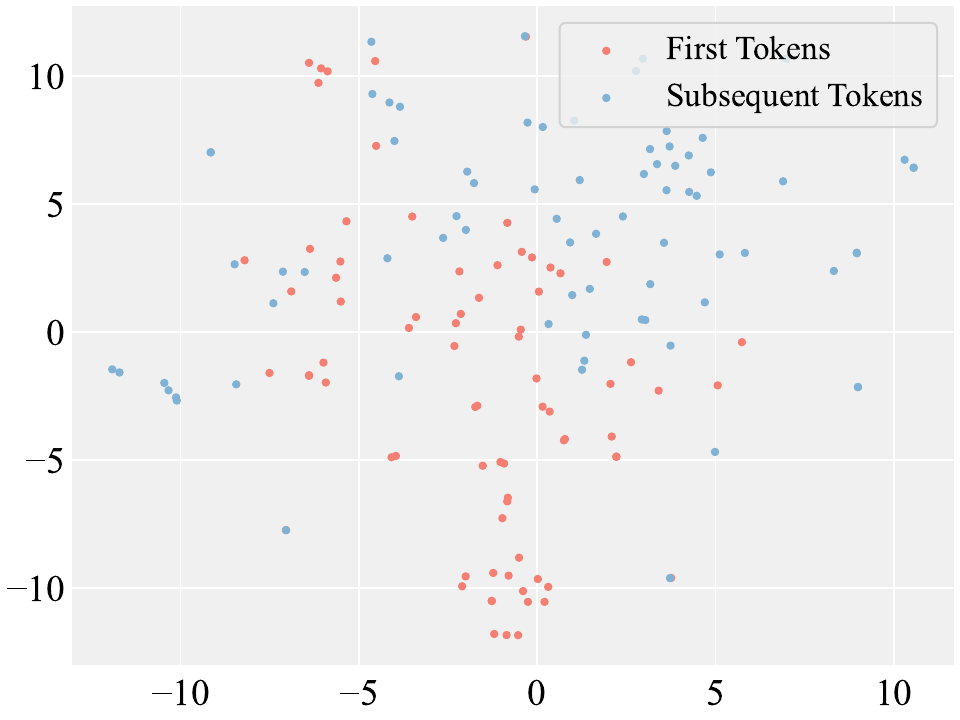}
    \captionsetup{skip=2pt}
    \caption{layer 10}
  \end{subfigure}
  \begin{subfigure}[t]{0.49\columnwidth}
    \includegraphics[width=\linewidth]{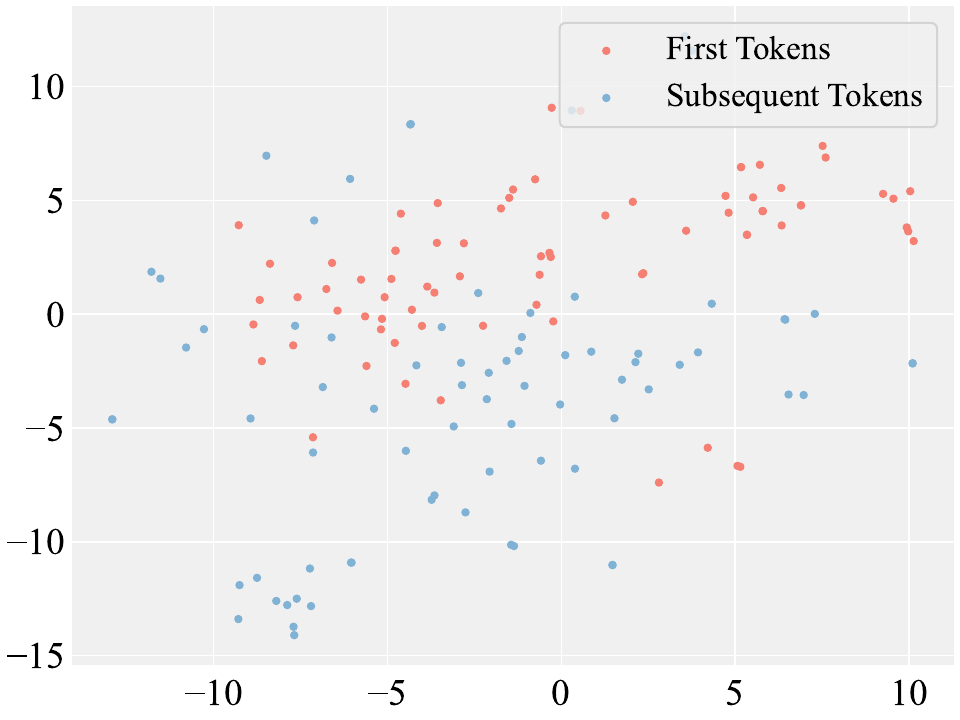}
    \captionsetup{skip=2pt}
    \caption{layer 15}
  \end{subfigure}

  \vspace{5pt}

  \begin{subfigure}[t]{0.49\columnwidth}
    \includegraphics[width=\linewidth]{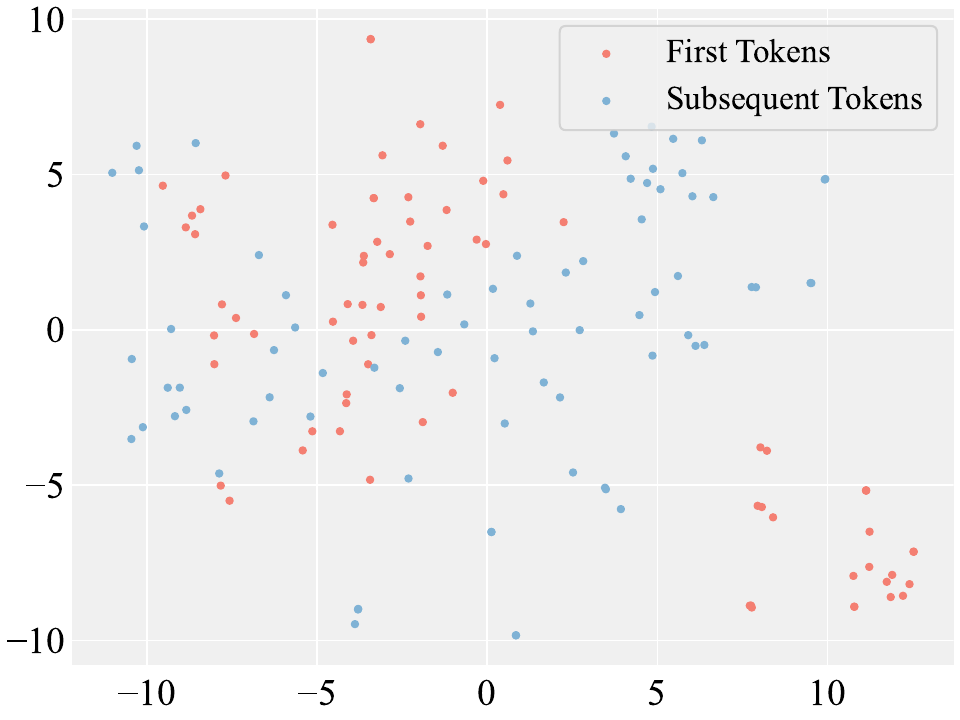}
    \captionsetup{skip=2pt}
    \caption{layer 20}
  \end{subfigure}
  \begin{subfigure}[t]{0.49\columnwidth}
    \includegraphics[width=\linewidth]{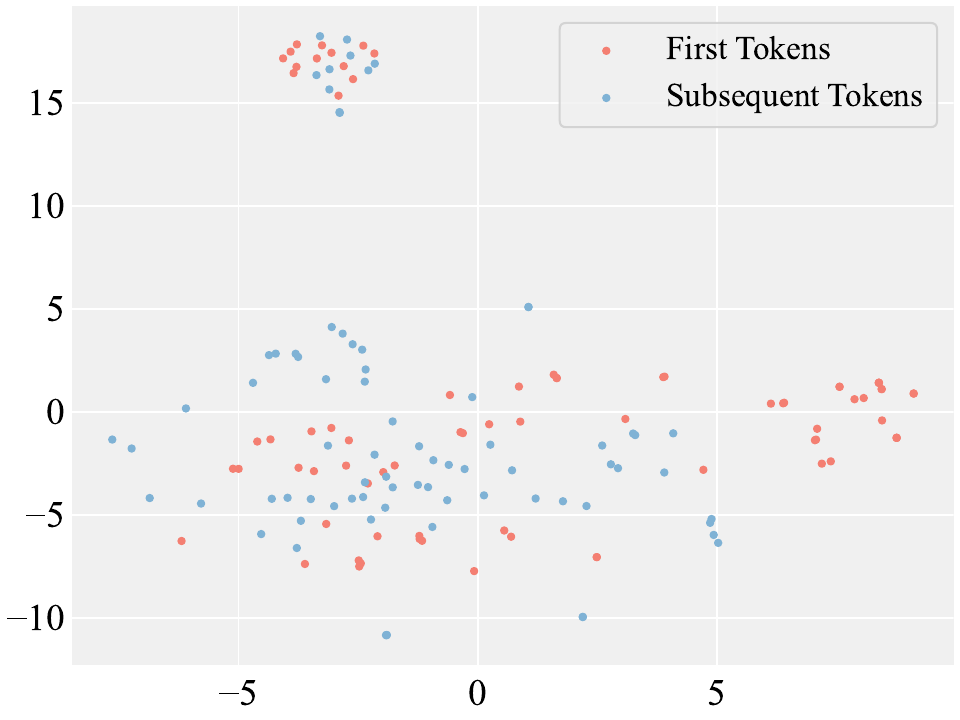}
    \captionsetup{skip=2pt}
    \caption{layer 23 (last layer)}
  \end{subfigure}

  \caption{t-SNE visualization of representations for first tokens and subsequent tokens across various layers in the encoder of T5-3B.}
  \label{fig_t5_label}
\end{figure}

\begin{figure*}
    \centering
    \includegraphics[width=\linewidth]{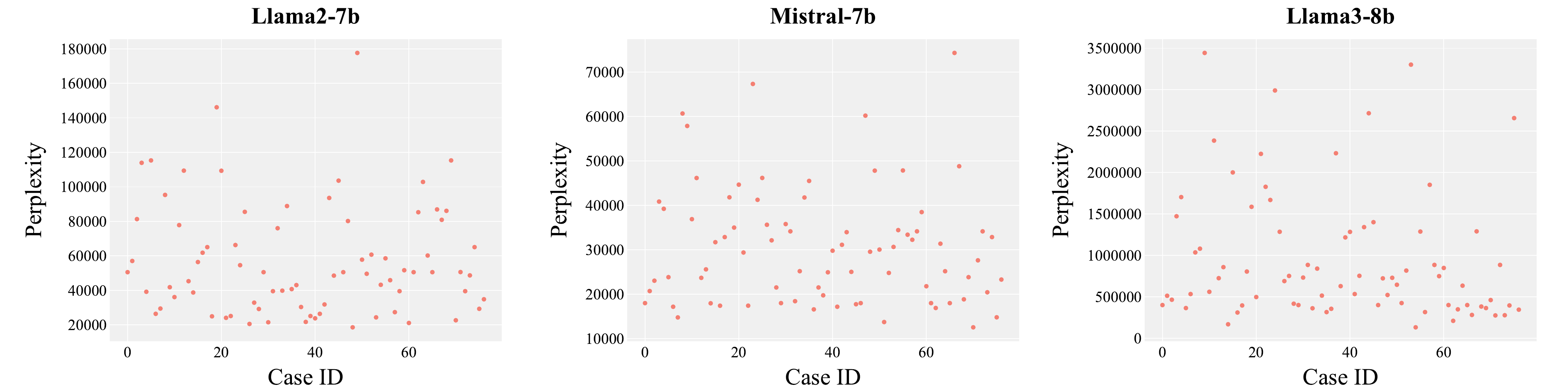}
    \caption{Scatter plot of perplexity for Llama2-7b, Mistral-7b, and Llama3-8b models edited by ROME, with each point representing a unique edit case in the collapse cases of GPT-2-XL and GPT-J. 
    ``Case ID'' refers to the index of each edit sample.}
    \label{fig_llama_del}
\end{figure*}

To validate that the absence of collapse in Llama2-7b, Mistral-7b, and Llama3-8b for the collapse cases of GPT-2-XL and GPT-J, is due to the addition of a prefix token, we manually removed the prepended token of these models, thereby positioning the unprefixed key $\bm{k}^{u}$ of the collapse cases as the first token of the input.
In this setting, we employed ROME to edit these three models on the collapse cases of GPT-2-XL and GPT-J.
The results presented in Figure~\ref{fig_llama_del} indicate that Llama2-7b, Mistral-7b, and Llama3-8b also succumb to collapse after editing.

\subsection{Representation of First Token in T5-3B}
\label{apd_t5}

The anomalous representation distribution of the first tokens in autoregressive models may be attributed to their inability to interact with subsequent tokens.
To verify it, we take the encoder-decoder model T5-3B as a counterexample and analyze the representation distribution of the first tokens in the collapse cases compared to an equal number (\num{77}) of subsequent tokens from the normal cases across various layers in its encoder.
The results in Figure~\ref{fig_t5_label} indicate that there is no significant difference between the representations of the first token and subsequent tokens, corroborating our hypothesis.

\subsection{Impact of Position Embedding}
\label{apd_position_embd}

In this section, we conducted experiments on GPT-2-XL, GPT-J, and Llama2-7b to investigate whether the anomalous distribution of the first token is attributable to its position embedding.
For Llama2-7b, we removed the special token \texttt{<s>} that the tokenizer additionally prepends at the beginning of the input to maintain consistency with GPT-2-XL and GPT-J.

\begin{table}[t]
    \centering
    \begin{adjustbox}{max width=0.48\textwidth} 
    \begin{tabular}{llrr}
        \toprule
        Model & Perplexity & Original & Second2First \\
        \midrule
        \multirow{3}{*}{GPT-2-XL} & min & \num{2177.82} & \num{1008.21} \\
         & avg & \num{19877.79} & \num{1397.87} \\
         & max & \num{179185.99} & \num{2153.86} \\
         \midrule
         \multirow{3}{*}{GPT-J} & min & \num{5094.73} & \num{8153.70} \\
         & avg & \num{28835.21} & \num{26978.14} \\
         & max & \num{85936.24} & \num{124982.41} \\
         \midrule
         \multirow{3}{*}{Llama2-7b} & min & \num{16279.75} & \num{17561.97} \\
         & avg & \num{67436.51} & \num{72692.50} \\
         & max & \num{206307.60} & \num{349577.58} \\
        \bottomrule
    \end{tabular}
    \end{adjustbox}
    \caption{The minimum, average, and maximum perplexity observed in collapse cases when utilizing the original position embeddings (Original) and when assigning the first token's position embedding as that of the second token (Second2First) for various LLMs.}
    \label{tab_sec2first}
\end{table}

For collapse cases where the subjects are the first tokens, we set the position embedding of the first token as that of the second token (Noted as Second2First).
The results presented in Table~\ref{tab_sec2first} indicate that this approach mitigates model collapse on GPT-2-XL, but it is completely ineffective on GPT-J and Llama2-7b.

\begin{table}[t]
    \centering
    \begin{adjustbox}{max width=0.48\textwidth} 
    \begin{tabular}{llrr}
        \toprule
        Model & Perplexity & Original & First2Second \\
        \midrule
        \multirow{3}{*}{GPT-2-XL} & min & \num{68.55} & \num{81.39} \\
         & avg & \num{68.81} & \num{39714.90} \\
         & max & \num{69.03} & \num{912001.20} \\
         \midrule
         \multirow{3}{*}{GPT-J} & min & \num{48.80} & \num{48.47} \\
         & avg & \num{49.03} & \num{48.68} \\
         & max & \num{49.50} & \num{49.48} \\
         \midrule
         \multirow{3}{*}{Llama2-7b} & min & \num{32.83} & \num{33.14} \\
         & avg & \num{33.32} & \num{2104.90} \\
         & max & \num{37.03} & \num{42154.10} \\
        \bottomrule
    \end{tabular}
    \end{adjustbox}
    \caption{The minimum, average, and maximum perplexity observed in normal cases when utilizing the original position embeddings (Original) and when assigning the second token's position embedding as that of the first token (First2Second) for various LLMs.}
    \label{tab_first2sec}
\end{table}

For normal cases where the subjects are the second tokens, we assign the position embedding of the second token as that of the first token (Noted as First2Second).
The results in Table~\ref{tab_first2sec} reveal that this change leads to partial model collapse in GPT-2-XL and Llama2-7b, but all edited models of GPT-J remain stable.

The results from the two aforementioned aspects suggest that position embedding may be a contributing factor to the abnormal representation of the first token, but it is not the sole factor.

\subsection{Collapse of First Token Representation}
\label{apd_collapse_1token}

From Figure~\ref{fig_dist} and Figure~\ref{fig_sentence}, we observed an unusual phenomenon that the collapse keys $\bm{k}^{u}$ (i.e., representations of the first tokens) appear to be more concentrated than the normal keys $\bm{k}^{u}$ (i.e., representations of the subsequent tokens).
To assess the degree of aggregation of the first tokens and subsequent tokens, we calculated the average distance of each element from the cluster center for both the first tokens and all the subsequent tokens, denoted as $D\left(F\right)$ and $D\left(S\right)$, correspondingly.
\begin{equation}
    D = \frac{1}{N} \sum_{i=1}^{N} \left\| \bm{e}_i - \frac{1}{N} \sum_{k=1}^{N} \bm{e}_k \right\|_2
\end{equation}
Here, $\bm{e}_i$ and $\bm{e}_k$ represent the embeddings of the $i$-th and $k$-th tokens, which are the outputs of the first MLP layer within the transformer block.

With this metric established, we computed the values within the edited layers of GPT-2-XL, yielding $D\left(F\right)$ being \num{0.578} and $D\left(S\right)$ being \num{13.895}. 
The result suggests a markedly higher concentration in the representations of the first tokens compared to those of subsequent tokens.
This observation raises a further question: \textit{Given that different first tokens have distinct embeddings when input into the transformer, why are their representations in the middle layers so closely concentrated?}

\begin{figure}[t]
  \centering

  \begin{subfigure}[t]{0.49\columnwidth}
    \includegraphics[width=\linewidth]{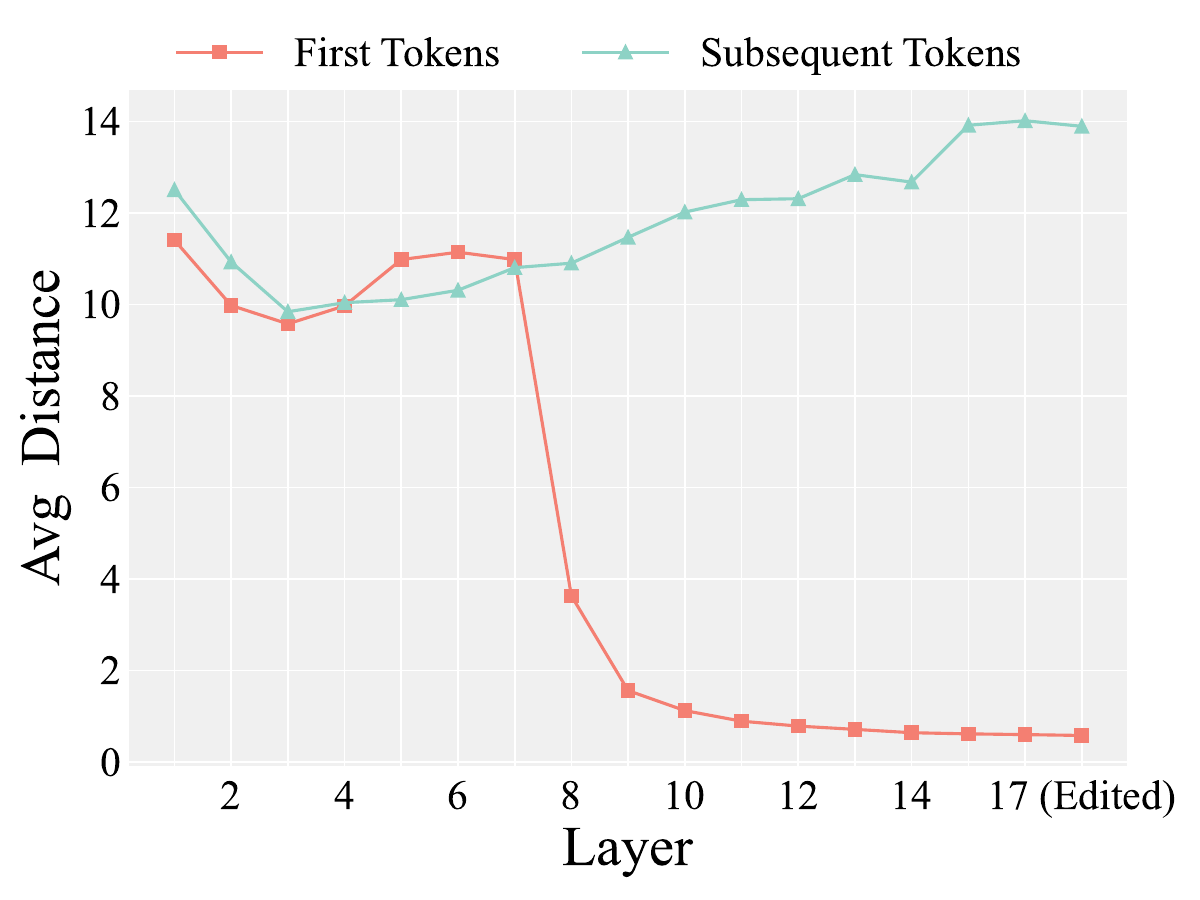}
    \caption{GPT-2-XL}
    \label{collapse_1token_gpt2}
  \end{subfigure}
  \hfill %
  \begin{subfigure}[t]{0.49\columnwidth}
    \includegraphics[width=\linewidth]{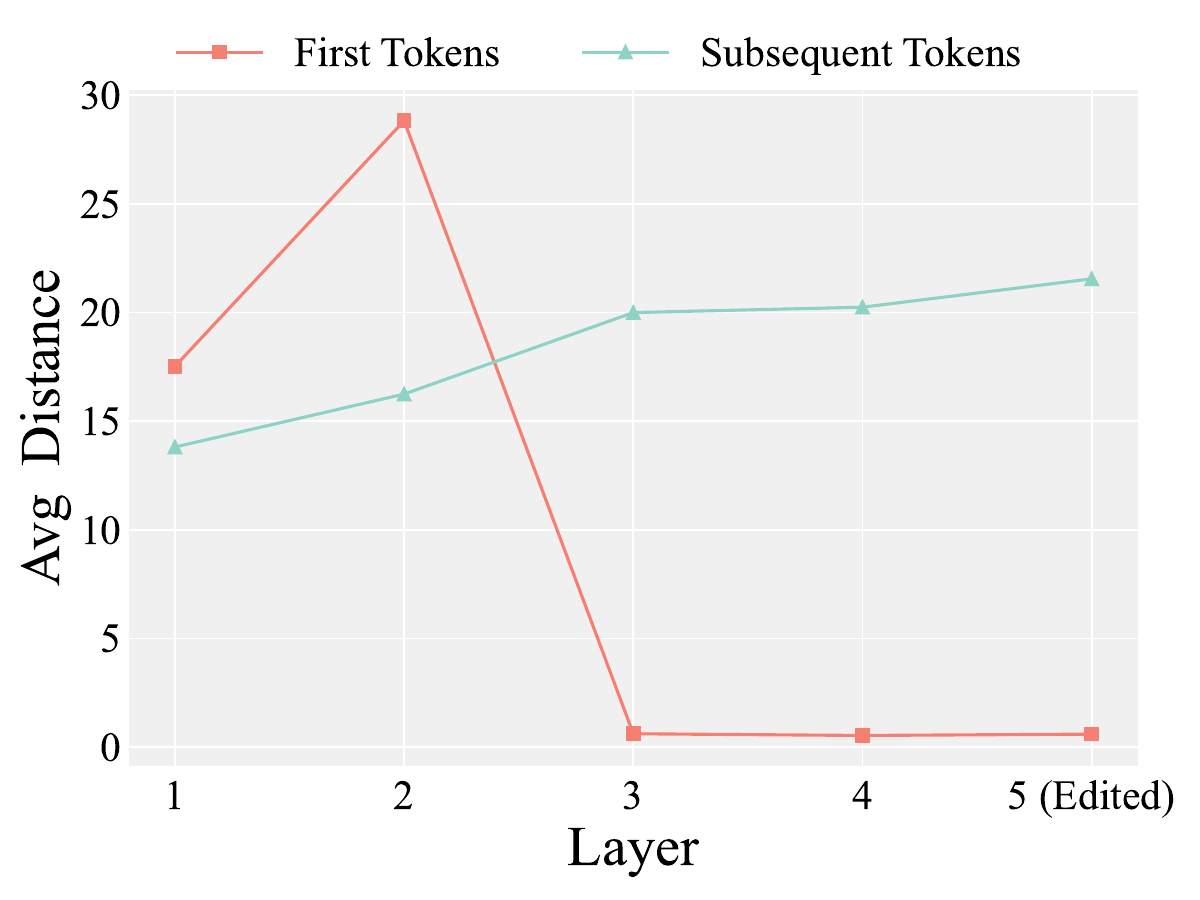}
    \caption{GPT-J}
    \label{collapse_1token_gptj}
  \end{subfigure}
  \caption{Average distances of each element from the cluster center for the first tokens and the subsequent tokens, across layers from the first layer to the edited layer in GPT-2-XL and GPT-J.}
  \label{fig_collapse_1token}
\end{figure}

To investigate this, we computed the distances $D\left(F\right)$ and $D\left(S\right)$ from the first layer to the edited layer (layer 17) in GPT-2-XL. 
As depicted in Figure~\ref{collapse_1token_gpt2}, prior to layer 8, $D\left(F\right)$ and $D\left(S\right)$ exhibit no significant divergence. 
However, post layer 8, the representations of the first tokens rapidly shrink.
The same phenomenon is also observed in GPT-J, as shown in Figure~\ref{collapse_1token_gptj}.
However, our experimental results indicate that such phenomenon does not appear on Llama2-7b, Mistral-7b, and Llama3-8b. Consequently, we decide not to delve further into this particular aspect.

The underlying causes of the first token's representation concentration in GPT-2-XL and GPT-J remain unclear. 
A potential factor, as explored in Appendix~\ref{apd_t5}, is that within autoregressive LLMs, the first token cannot interact with subsequent tokens. 
Continuous self-interaction may lead to the contraction of its representation.
Since this phenomenon is not related to the model collapse during editing examined in this paper, it has been remained for future research.

\end{document}